\pdfoutput=1

\documentclass[11pt]{article}

\usepackage{emnlp2021}

\usepackage{times}
\usepackage{latexsym}

\usepackage[T1]{fontenc}

\usepackage[utf8]{inputenc}

\usepackage{microtype}

\usepackage[export]{adjustbox}
\usepackage{graphicx}
\usepackage{algorithm}
\usepackage{verbatim}
\usepackage{algorithmic}
\usepackage{amsmath,amssymb,amsfonts}
\usepackage{caption}
\usepackage{subcaption}
\usepackage{graphicx}

\title{Phonetic Word Embeddings}

 \author{Rahul Sharma\thanks{ \hspace{4pt}equal contribution}       , Kunal Dhawan\footnotemark[1]      ,      Balakrishna Pailla \\
         Reliance Jio AI-CoE \\}

\begin{document}
\maketitle
\begin{abstract}
This work presents a novel methodology for calculating the phonetic similarity between words taking motivation from the human perception of sounds. This metric is employed to learn a continuous vector embedding space that groups similar sounding words together and can be used for various downstream computational phonology tasks. The efficacy of the method is presented for two different languages (English, Hindi) and performance gains over previous reported works are discussed on established tests for predicting phonetic similarity. To address limited benchmarking mechanisms in this field, we also introduce a heterographic pun dataset based evaluation methodology to compare the effectiveness of acoustic similarity algorithms. Further, a visualization of the embedding space is presented with a discussion on the various possible use-cases of this novel algorithm. An open-source implementation is also shared to aid reproducibility and enable adoption in related tasks.
\end{abstract}

\section{Introduction}
\label{sec:intro}
Word embeddings have become an indispensable element of any Natural Language Processing (NLP) pipeline. This technique transforms written words to a higher dimensional subspace usually on the basis of semantics, providing algorithms the benefits of vector spaces like addition and projection while dealing with words. They are currently being used for diverse tasks like question answering systems \citep{zhou2015learning}, sentiment analysis \citep{giatsoglou2017sentiment}, speech recognition \citep{bengio2014word}, and many more.
These embedding spaces are usually learned on the basis of the meaning of the word, as highlighted in the seminal work of word2vec \citep{mikolov2013distributed}. 

In this work, we explored a parallel ideology for learning word embedding spaces - based on the phonetic signature of the words. This methodology places two words in the embedding space closer if they sound similar. Such an embedding finds application in various tasks like keyword detection \citep{sacchi2019open} and poetry generation \citep{parrish2017poetic}. In \citep{parrish2017poetic}, the author clearly defines the concept of phonetic similarity:  
\begin{quote}
    Alliteration, assonance, consonance and rhyme are all figures of speech deployed in literary writing in order to create patterns of sound, whether to focus attention or to evoke synthetic associations. These figures are all essentially techniques for introducing \emph{phonetic similarity} into a text, whether that similarity is local to a single word, or distributed across larger stretches of text.
\end{quote}
\citep{parrish2017poetic} introduces a strategy of using phonetic feature bi-grams for computing vector representation of a word. This combines the notion that phonemes can be differentiated from each other using a limited set of properties (Distinctive feature theory \citep{chomsky1968sound}) and that articulation of speech is not discrete, thus acoustic features are affected by the phonemes preceding and following it \citep{browman1992articulatory}. We build upon these ideas and further try to incorporate the way in which humans perceive speech and acoustic similarity. This ideology led us to propose a novel phonetic similarity metric which is presented in detail in this paper.

The contributions of this paper are as follows: (i) algorithm for similarity between two phonemes based on well defined phonetic features; (ii) an edit distance \cite{wagner1974string} based approach for computing the acoustic similarity between two words; (iii) a novel non-diagonal penalty for matching the phonetic sequence of two words; (iv) a novel end of the word phone weightage factor that captures the perceptual sound similarity ability of humans; and (v) results are reported for two languages, English and Hindi. The results for English comfortably outperform previous best-reported results and this is the first instance that such results are reported for Hindi, to the best of our knowledge.

The remainder of the paper is organized as follows- We present a literature review in Section~\ref{sec:litreview}, Section~\ref{sec:approach} discusses the proposed approach in detail, the experiments performed and results are discussed in Section~\ref{sec:result} and finally the work is concluded in Section~\ref{sec:conclusion}.

The source code, for the proposed method alongside experiments, has been released publicly and can be accessed at \href{https://github.com/rahulsrma26/phonetic-word-embedding}{https://github.com/rahulsrma26/phonetic-word-embedding} 

\section{Related work}
\label{sec:litreview}
 In literature, there have been multiple works that employ phonetic similarity techniques for various downstream tasks like studying language similarities, diachronic language change, comparing phonetic similarity with perceptual similarity, and so on. This diverse usage acts as a strong motivation for us to come up with a more accurate and robust phonetic similarity-based word embedding algorithm.

\citep{bradlow2010perceptual} explores methods representing languages in a perceptual similarity space based on their overall phonetic similarity. It aims to use this phonetic and phonological likeness of any two languages to study patterns of cross-language and second-language speech perception and production. It is important to note that their focus is not on words in a language but rather the usage of the similarity space to study the relationship between languages. \citep{mielke2012phonetically} talks about quantifying phonetic similarity and how phonological notions of similarity are different from articulatory, acoustic, and perceptual similarity. The proposed metric, which is based on phonetic features, was compared with measures for phonological similarity, which are calculated by counting the co-occurrences of pairs of sounds in the same phonologically active classes. In the experiments conducted by \citep{vitz1973predicting}, small groups of L1 American English speakers were asked to score the phonetic similarity of a “standard” word with a list of comparison words on a scale from 0 (no similarity) to 4 (extremely similar). Several experiments were performed with different standard words and comparison words. Vitz and Winkler compared the elicited scores to the results from their own procedure for determining phonetic similarity (termed “Predicted Phonemic Distance,” shortened here as PPD). \citep{parrish2017poetic} proposes a novel technique of using phonetic feature bi-grams for computing vector representation of a word. The technique was shown to perform better than \citep{vitz1973predicting}'s PPD approach in some cases. The author works with version $0.7$b of the CMU Pronouncing Dictionary \cite{cmudict}, which consists of $133852$ entries, appends a token 'BEG' to beginning \& 'END' to the end of the phonetic transcription of all the words and calculates interleaved feature bi-grams for each of them. Using the fact that there can be 949 unique interleaved feature bi-grams across the entire dataset, the author represent the features extracted for all the words as a matrix of size $133852$ (number of words in the dictionary) $\times$ $949$ (number of unique features). PCA is applied to the above matrix to get $50$ dimensional embedding for each of the words in the dataset, which we refer to as PSSVec in this paper, short for Poetic Sound Similarity Vector. To capture the similarity of words regardless of the ordering of phonemes, the author further computes the features for the reverse order of the phonemes and appends that to the original feature of the word. Thus, a drawback of this approach is that embedding is not dependent of the position of the phoneme in the word, for example, the embedding for phoneme sequences 'AY P N OW' and 'OW N P AY' would turn out to be same, which is counter-intuitive as they don't sound similar. Hence, we postulate that the position of phonemes also plays a major role for two words to sound similar (and thus be closer to each other in the embedding space) and we address this issue in our proposed similarity algorithm.

\begin{figure*}[]
  \centering
    \captionof{table}{Complete feature description for English consonants (* denotes multiple occupancy of the given phoneme)}
  \includegraphics[width=\linewidth]{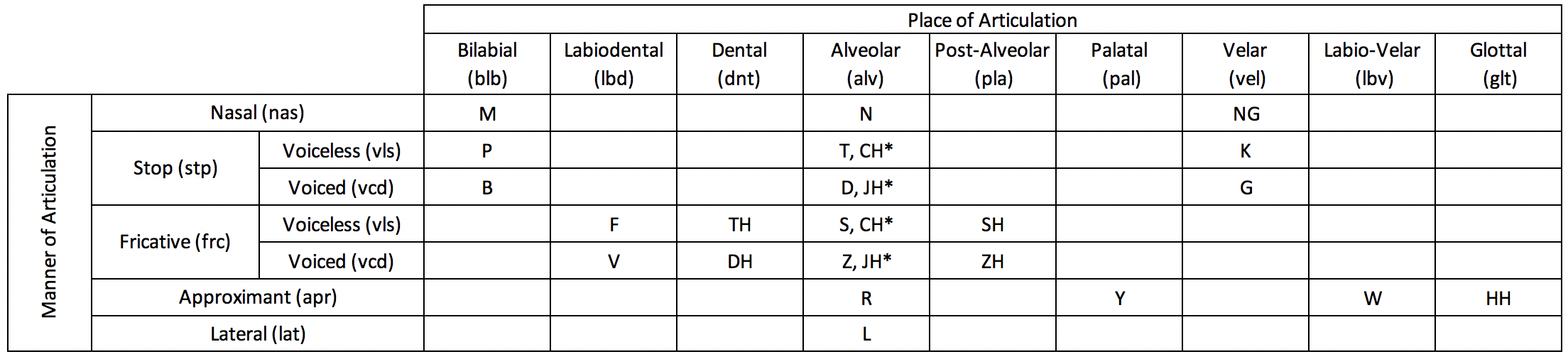} \vspace{2mm}
  \label{fig:english_consonants}
\end{figure*}

\begin{figure*}[]
  \centering
    \captionof{table}{Complete feature description for English vowels (* denotes multiple occupancy of the given phoneme)}
  \includegraphics[scale=0.46]{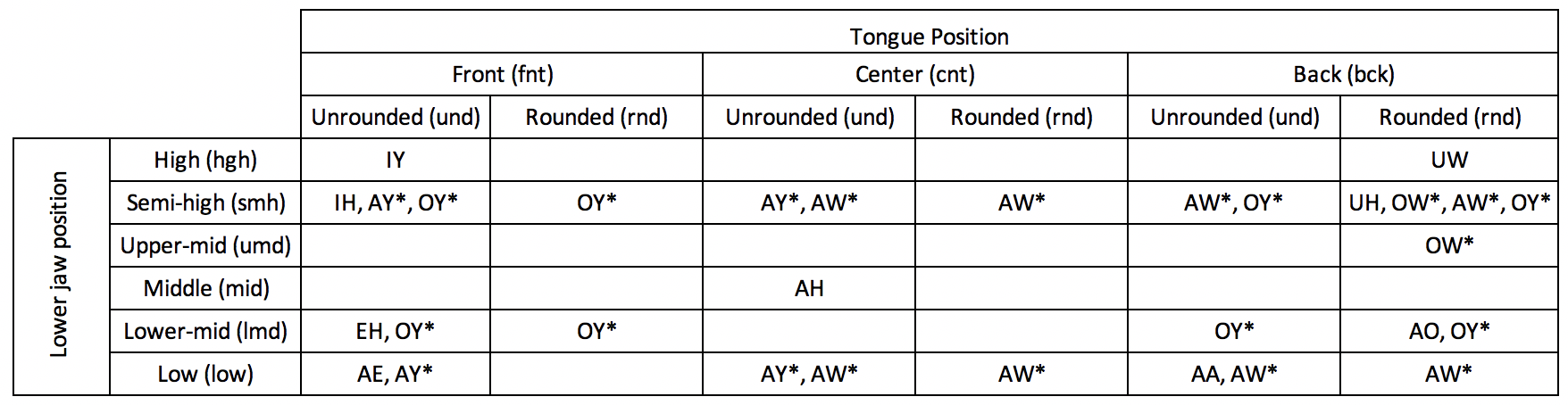} \vspace{2mm}
  \label{fig:english_vowels}
\end{figure*}

\begin{figure*}[]
  \centering
    \captionof{table}{Complete feature description for Hindi consonants}
  \includegraphics[scale=0.3]{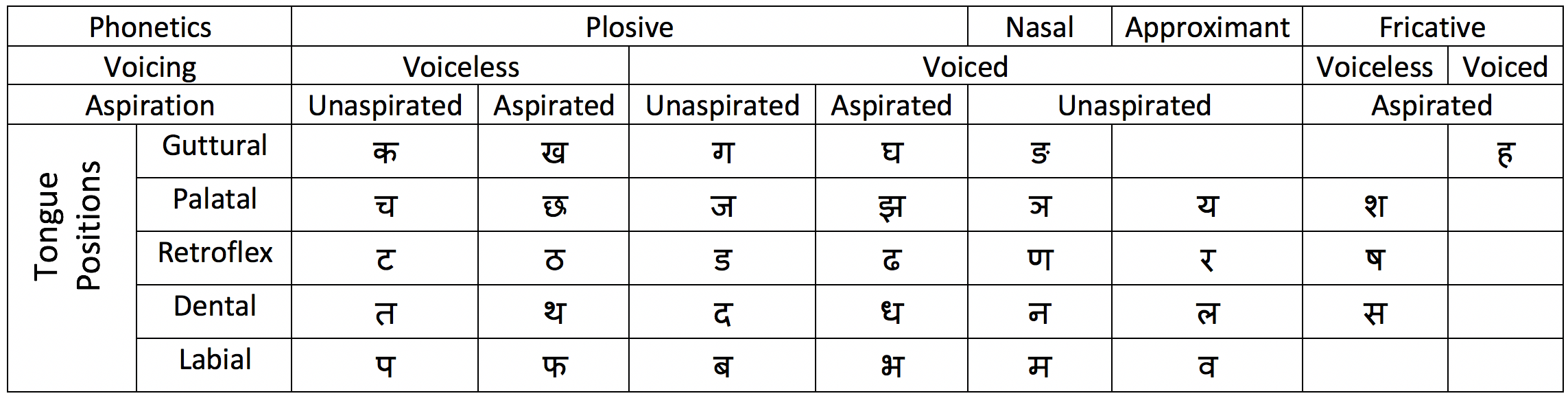}  \vspace{2mm}
  \label{fig:hindi_consonants}
\end{figure*}

\begin{figure*}[]
  \centering
    \captionof{table}{Complete feature description for Hindi vowels}
  \includegraphics[scale=0.3]{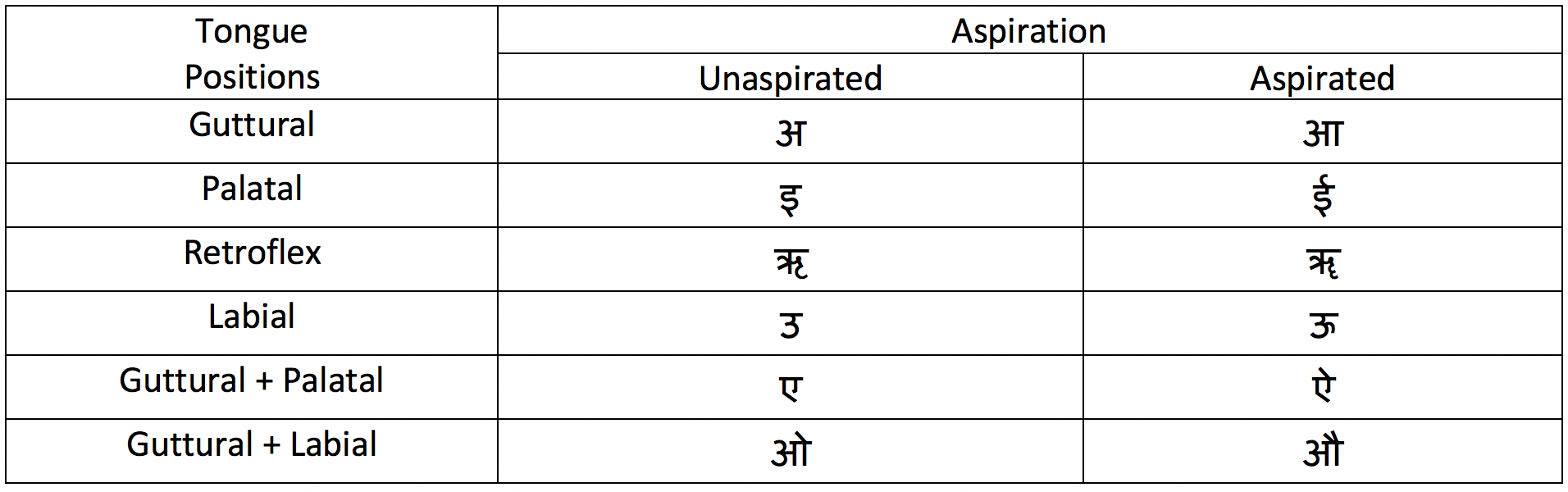}  \vspace{2mm}
  \label{fig:hindi_vowels}
\end{figure*}

\section{Proposed Approach}
\label{sec:approach}

It has been well established by linguists that some phonemes are more similar than others depending upon the variations in articulatory positions involved while producing them \cite{chomsky1968sound}, and these variations are captured by phoneme features. Based on these features phonemes can be categorized into groups. 
Our method assumes that each phoneme can be mapped to a set of corresponding phonetic features. For English, we use the mapping suggested in the specification for X-SAMPA \cite{Kirshenbaum2001}, adapted to the Arpabet transcription scheme. The mapping of English phonemes to features is shown in table \ref{fig:english_consonants} and \ref{fig:english_vowels}.

After Mandarin, Spanish and English, Hindi is the most natively spoken language in the world, almost spoken by 260 million people according to \cite{widelySpoken}.
For Hindi, we use the mapping suggested by \cite{malviya2016structural} and added some of the missing ones using IPA phonetic features \cite{mortensen2016panphon}.
The mapping of Hindi phonemes to features used by us is shown in tables \ref{fig:hindi_consonants} and \ref{fig:hindi_vowels}.

\subsection{Phoneme Similarity}

The phonetic distance (Similarity) $S$ between a pair of phonemes $P_a$ and $P_b$ is measured by using the feature set of two phonemes and computing Jaccard similarity. 
\begin{equation}
S(P_a, P_b) = \frac{|F(P_a)\cap F(P_b)|}{|F(P_a)\cup F(P_b)|}
\end{equation}

Here, $F(P)$ is the set of all the feature of phone $P$. For example, $F(R) = \{apr, alv\}$ (from table \ref{fig:english_consonants}). We can extend this logic to bi-grams as well. To compute the feature set of bi-gram $(P_{a1},P_{a2})$, we will just take the union of the features for both of the phones.
\begin{equation}
F(P_{a1}, P_{a2}) = F(P_{a1})\cup F(P_{a2})
\end{equation}

And Similarity $S$ between pairs of bi-grams $(P_{a1}, P_{a2})$ and $(P_{b1}, P_{b2})$, can be calculated as:
\begin{equation}
\label{eq:bigram}
\begin{split}
S((P_{a1}, P_{a2}),(P_{b1}, P_{b2})) = \hspace{30mm}  \\
\frac{|F(P_{a1}, P_{a2})\cap F(P_{b1}, P_{b2})|}{|F(P_{a1}, P_{a2})\cup F(P_{b1}, P_{b2})|}
\end{split}
\end{equation}

Vowels in speech are in general longer than consonants \cite{umeda1975vowel} \cite{umeda1977consonant} and further the perception of similarity is heavily dependent upon vowels \cite{raphael1972preceding}. This is the reason that if two bi-grams are ending with the same vowel then they sound almost the same despite them starting with different consonants. 

We can weight our similarity score to reflect this idea.
\begin{equation}
\label{eq:vowel_weight}
\begin{split}
S_v((P_{a1}, P_{a2}),(P_{b1}, P_{b2})) = \quad \quad \quad \quad \quad \quad \quad \quad \\
\begin{cases}
    \sqrt{S((P_{a1}, P_{a2}),(P_{b1}, P_{b2}))}, \\
    \quad \quad \quad \text{if } P_{a2} \text{ is a vowel and } P_{a2} = P_{b2}\\
    \quad \\
    S((P_{a1}, P_{a2}),(P_{b1}, P_{b2}))^2, \quad \text{otherwise}
\end{cases}
\end{split}
\end{equation}

Here, $S_v$ is the vowel-weighted similarity for the bi-grams $(P_{a1}, P_{a2})$ and $(P_{b1}, P_{b2})$.

\subsection{Word Similarity}

We are proposing a novel method to compare words that take the phoneme similarity as well as their sequence into account. The symmetric similarity distance between word $a$, given as the phoneme sequence ${a_1, a_2, a_3 \ldots a_n}$, and word $b$, given as the phoneme sequence ${b_1, b_2, b_3 \ldots b_m}$ is given by $d_{nm}$, defined by the recurrence:




\begin{equation*}
\label{eq:similarity_recurrence}
d_{1,1} = S(a_1, b_1)
\end{equation*}
\begin{equation*}
d_{i,1} = d_{i - 1, 1} + S(a_i, b_1) \qquad \text{for } 2 \leq i \leq n
\end{equation*}

\begin{equation*}
d_{1,j} = d_{1, j - 1} + S(a_1, b_j) \qquad \text{for } 2 \leq j \leq m 
\end{equation*}

\begin{equation}
\begin{array}{ll}
d_{i,j} = 
\begin{cases}
    S(a_i,b_j) + d_{i-1,j-1}, \text{if } S(a_i,b_j) = 1\\
    \begin{array}{ll}
        S(a_i,b_j)/p \\
        \quad + 
        min\left\{
            \begin{array}{ll}
                d_{i-1, j}\\
                d_{i, j-1}
            \end{array}
        \right.
    \end{array}
    ,\text{otherwise}
\end{cases} \\
\hspace{30mm} \text{for } 2 \leq i \leq n, 2 \leq j \leq m
\end{array}
\end{equation}

Here, $p$ is the non-diagonal penalty. As we are calculating scores for all possible paths, this allows us to discourage the algorithm from taking convoluted non-diagonal paths, which will otherwise amass better scores as they are longer. The word similarity $W_S$ between the words $a$ and $b$ can be calculated as:
\begin{equation}
    W_S(a, b) = d_{n,m} / max(n, m)
\end{equation}
By setting non-diagonal penalty $p \geq 2$, we can ensure that word similarity $W_S$ will be in the range $0 \leq W_S \leq 1$.

Further, the score can be calculate for a bi-gram sequence instead of a uni-gram sequence by using equation \ref{eq:bigram}. In case of bi-grams, we  inserte two new tokens: 'BEG' in the beginning and 'END' in the end of the phoneme sequence. For word $a$, defined by the original sequence ${a_1, a_2, a_3 \ldots a_n}$, the bi-gram sequence will be $(BEG,a_1), (a_1,a_2), (a_2,a_3), (a_3,a_4) \ldots \\ (a_{n-1},a_n), (a_n,END)$. Here, 'BEG' and 'END' are the dummy phones which are mapped to the single dummy phonetic features 'beg' and 'end' respectively.

We can also utilize the benefits of vowel weight concept that we established in equation \ref{eq:vowel_weight}. The only change required in the equation \ref{eq:similarity_recurrence} is to use $S_v$ instead of $S$ in the calculation of symmetric similarity distance $d_{n,m}$. We denote the word similarity obtained via this method as $W^V_S$ (vowel weighted word similarity) in this paper.

\subsection{Embedding Calculation}

Using the algorithm mentioned in the previous section, we can obtain similarities between any two words, as long as we have their phonemic breakdown. In this section, we highlight the approach followed to build an embedding space for the words. Word embeddings offer many advantages like faster computation and independent handling in addition to the other benefits of vector spaces like addition, subtraction, and projection.

For a $k$ word dictionary, we can obtain the similarity matrix $M \in \mathbb{R}^{k * k}$ by computing the word similarity between all the pairs of word. 
\begin{equation}
\label{eq:SimilarityMatrix}
    M_{i,j} = W^V_S(word_i, word_j)
\end{equation}
Since this similarity matrix $M$ will be a symmetric non-negative matrix, one can use Non-negative Matrix Factorization \cite{yang2008non} or a similar technique to obtain the $d$ dimension word embedding $V \in \mathbb{R}^{k * d}$. Using Stochastic Gradient Descent(SGD) \cite{koren2009matrix} based method we can learn the embedding matrix $V$ by minimizing:
\begin{equation}
\label{eq:embedding}
    || M - V.V^T ||^2
\end{equation}

We choose SGD based method as matrix factorization and other such techniques become inefficient due to the memory requirements as the number of words $k$ increases. For example to fit the similarity matrix $M$ in memory for $k$ words the space required will be in the order of $O(k^2)$. But using SGD based method we don't need to pre-calculate the $M$ matrix, we can just calculate the required values by using equation \ref{eq:SimilarityMatrix} on the fly.



\begin{figure*}
\centering
\begin{minipage}{.47\textwidth}
  \includegraphics[width=\linewidth, left]{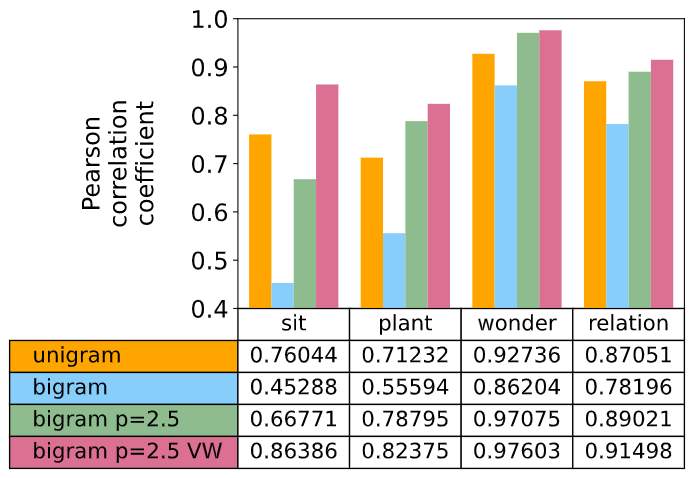}
  \captionof{figure}{Correlation of proposed methods with Human Survey}
  \label{fig:r_unigram}
\end{minipage}%
\qquad
\begin{minipage}{.47\textwidth}
  \includegraphics[width=\linewidth, right]{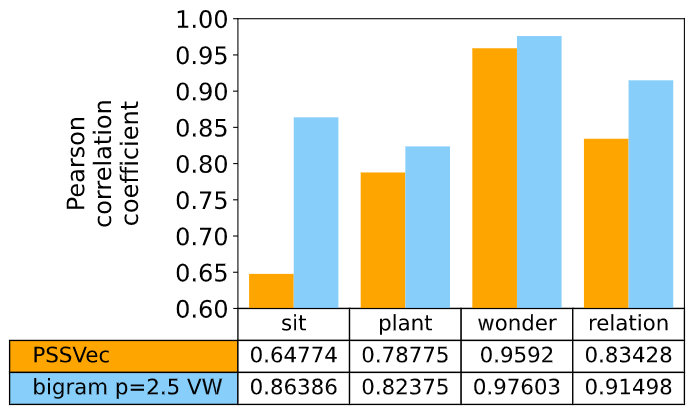}
  \captionof{figure}{Correlation of existing work and our method with Human Survey}
  \label{fig:r_bigram}
\end{minipage}
\end{figure*}


\begin{figure*}
\centering
\begin{minipage}{.47\textwidth}
  \includegraphics[width=\linewidth, left]{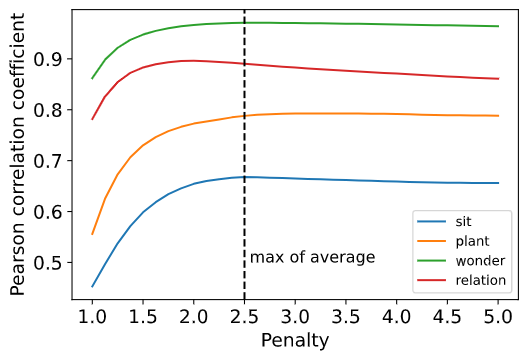}
  \captionof{figure}{Effect of the penalty on correlation}
  \label{fig:r_penalty}
\end{minipage}%
\qquad
\begin{minipage}{.47\textwidth}
  \includegraphics[width=\linewidth, right]{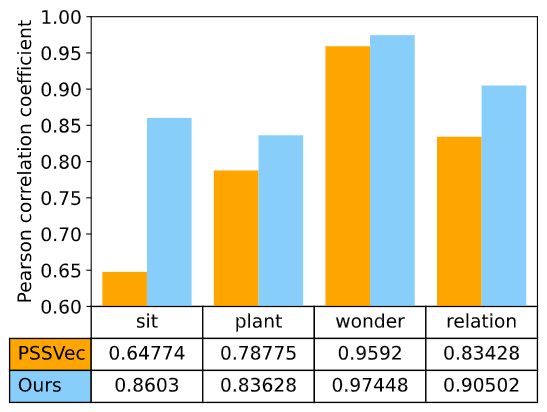}
  \captionof{figure}{Correlation of embeddings with Human Survey}
  \label{fig:r_embedding}
\end{minipage}
\end{figure*}


\begin{figure*}[]
  \centering
    \caption{Visualizing the acoustic embedding vectors of selected English words}
  \includegraphics[scale=0.35]{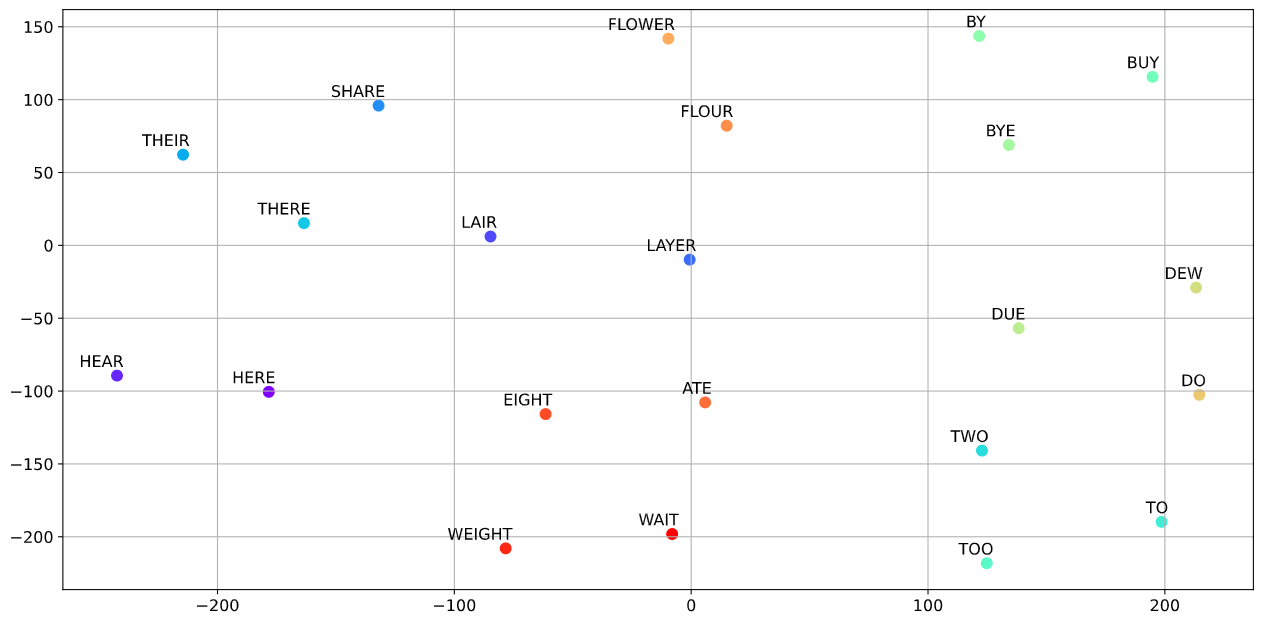} 
  \label{fig:visualization_english}
\end{figure*}

\begin{figure*}[]
  \centering
    \caption{Visualizing the acoustic embedding vectors of selected Hindi words}
  \includegraphics[scale=0.35]{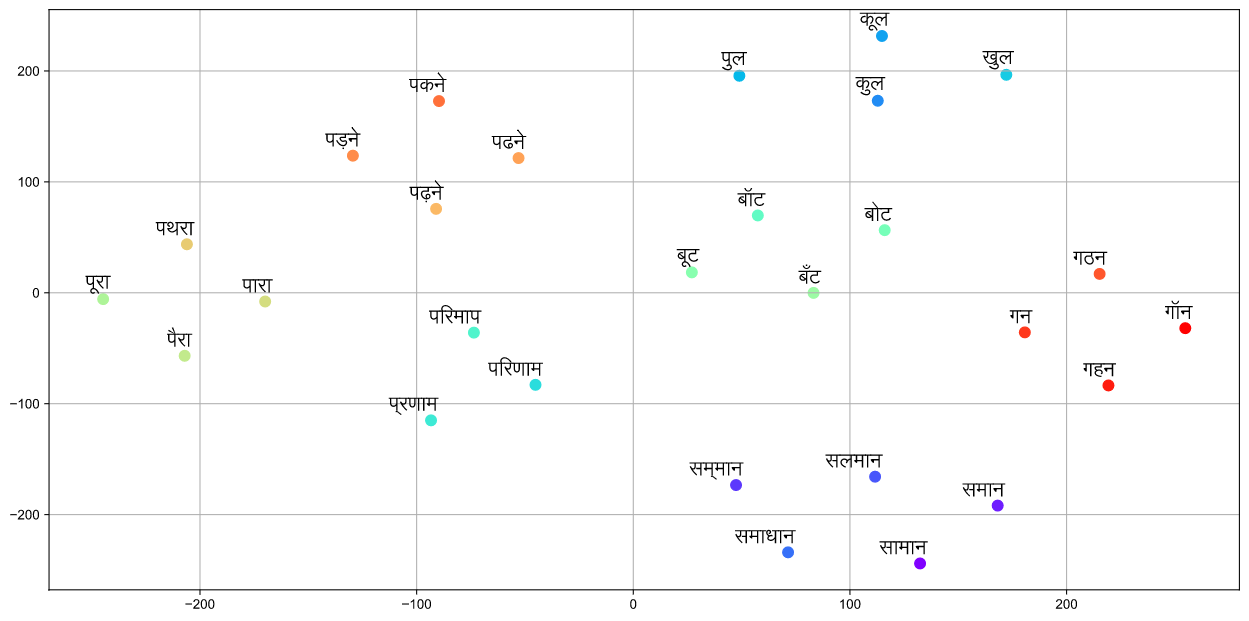}
  \label{fig:visualization_hindi}
\end{figure*}


\section{Experiments and Results}
\label{sec:result}

\subsection{Experiments}
We implemented the memoized version of the algorithm using dynamic programming. Given two words $a$ and $b$ with phoneme sequences ${a=a_1, a_2, a_3 \ldots a_n}$ and ${b=b_1, b_2, b_3 \ldots b_m}$, the complexity of the algorithm is $O(n.m)$. The algorithm is implemented in Python and CMU Pronouncing Dictionary 0.7b \cite{cmudict} is used to obtain the phoneme sequence for a given word.

As described below, for comparison of our proposed approach to existing work, we calculated the correlation between human scores to the scores given by PSSVec and our method for all the standard words as outlined in \citep{vitz1973predicting}. 

First we use our algorithm $W_S$ on uni-grams and then bi-grams. For bi-grams we have used 3 variations: one without non-diagonal penalty (i.e. $p=1$), one with $p=2.5$ and one with the vowel weights ($W^V_S$). We compared the correlation between methods and the human survey results \cite{vitz1973predicting}. We scaled the human survey results from $0$-$4$ to $0$-$1$. As seen in figure \ref{fig:r_unigram}, giving preference to the matching elements by setting $p=2.5$ increased the score. Effect of the non-diagonal penalty $p$ can be seen in figure \ref{fig:r_penalty}. This also shows that giving higher weights to similar ending vowels further increased the performance.

After that, we compared our method (bi-gram with $W^V_S$ and $p=2.5$) with the existing work. As we can see from the figure \ref{fig:r_bigram}, our method out performs the existing work in all 4 cases.

\subsection{Embedding Space Calculation}
As the vowel weighted bi-gram based method $W^V_S$ with non-diagonal penalty $p=2.5$ gave us the best performance, we use this to learn embeddings by solving equation \ref{eq:embedding}. Tensorflow v2.1 \cite{abadi2016tensorflow} was used for implementation of our matrix factorization code..

As python implementation is slow, we have also implemented our algorithm in C++ and used it in Tensorflow (for on-fly batch calculation) by exposing it as a python module using Cython \cite{seljebotn2009fast}. In single-thread comparison we got an average speed-up around 300x from the C++ implementation over the Python implementation. The implementation benefits from the compiler intrinsics for bit-count \cite{mula2018faster} which speedup the Jaccard similarity computation. The results were obtained on Intel(R) Xeon(R) CPU E5-2690 v4 @ 2.60GHz 440GB RAM server machine running on Ubuntu 18.04. To compare the running performance, we use the time taken to compute the similarity between a given word to every word in the dictionary. The experiments were repeated 5 times and their average was taken for the studies.


Similar to PSSVec we also used 50 dimensional embeddings for a fair comparison. The English embeddings are trained on $133859$ words from CMU 0.7b dictionary \cite{cmudict}. Figure \ref{fig:r_embedding} shows the comparison between PSSVec and our embeddings.

For Hindi language, we use \cite{kunchukuttan2020indicnlp} dataset and train on $22877$ words.

We can also perform sound analogies (which reveal the phonetic relationships between words) on our obtained word embeddings using vector arithmetic to showcase the efficacy of the learned embedding space. Let's assume that $V(W)$ represents the learned embedding vector for the word $W$. If $W_a$ is related to $W_b$ via the same relation with which $W_c$ is related to $W_d$ ($W_a : W_b :: W_c : W_d $), given the embedding of words $W_a, W_b, W_c$, we can obtain $W_d$ as:
\begin{equation}
    W_d = N(V(W_b) - V(W_a) + V(W_c))
\end{equation}

where function $N(v)$ returns the nearest word from the vector $v$ in the learnt embedding space. The results for some pairs are documented in the following tables:

\begin{center}
  \includegraphics[width=\linewidth]{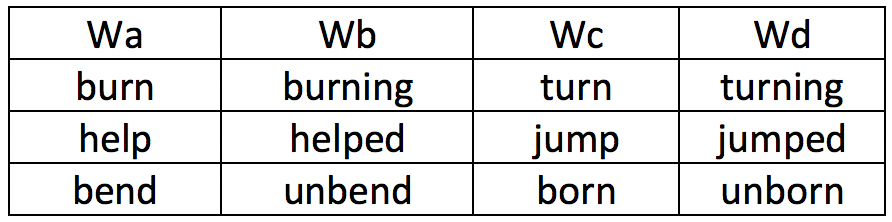}
  \captionof{table}{Sound analogies for English}
  \label{fig:r_english_calc}
\end{center}

\begin{center}
  \includegraphics[width=\linewidth]{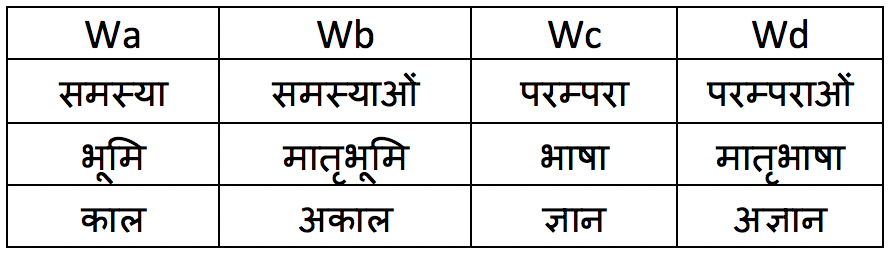}
  \captionof{table}{Sound analogies for Hindi}
  \label{fig:r_hindi_calc}
\end{center}

To aid visual understanding of this embedding space, we selected a few words each from English \& Hindi languages and projected their embeddings to a 2-dimensional space using t-distributed stochastic neighbor embedding \citep{maaten2008visualizing}. The projected vectors are presented as a 2D plot in figure \ref{fig:visualization_english} for English and figure \ref{fig:visualization_hindi} for Hindi. We clearly note that similar sounding words occur together in the plot, hence validating our hypothesis that our proposed embedding space is able to capture the phonetic similarity of words.

\subsection{Pun Based Evaluation}

A pun is a form of wordplay which exploits the different possible meanings of a word or the fact that there are words that sound alike but have different meanings, for an intended humorous or rhetorical effect \cite{miller2017semeval}. The first category where a pun exploits the different possible meanings of a word is called Homographic pun and the second category that banks on similar sounding words is called Heterographic pun. Following is an example of a Heterographic pun:

\begin{center}
I Renamed my iPod The Titanic, 

so when I plug it in, it says

“The Titanic is syncing.”
\end{center}

Heterographic puns are based on similar sounding words and thus act as a perfect benchmarking dataset for acoustic embedding algorithms. For showing the efficacy of our approach, we have used the 3rd subtask (heterographic pun detection) of the SemEval-2017 Task 7 dataset \cite{miller2017semeval}. This test set contains 1098 hetrographic puns word pairs like \textit{syncing} and \textit{sinking}. After taking all word pairs that are present in the CMU Pronouncing Dictionary \cite{cmudict} and removing duplicate entries, we are left with 778 word-pairs. Ideally, for a good acoustic embedding algorithm, we would expect the distribution of cosine similarity of these word pairs to be a sharp Gaussian with values $\rightarrow$ 1.

To further intuitively highlight the efficacy of our approach, for the 778 word-pairs we present in table \ref{fig:r_pun_diff} the difference of the cosine similarity scores obtained from our algorithm and PSSVec (sorted by the difference). As we can see, though the words \textit{mutter} and \textit{mother} are relatively similarly sounding, PSSVec assigns a negative similarity. In comparison, our method is more robust, correctly capturing the acoustic similarity between words.

\begin{center}
  \captionof{table}{Difference of scores}
  \label{fig:r_pun_diff}
 \begin{tabular}{c c c c c} 
 \hline
 Word1 & Word2 & PSSVec & Ours & Diff \\ [0.5ex] 
\hline\hline
 mutter &  mother & -0.0123 & 0.8993 &  0.9117 \\
\hline
 loin &   learn & -0.0885 & 0.8119 &  0.9005 \\
\hline
 truffle & trouble &  0.1365 & 0.9629 &  0.8264 \\
\hline
 soul & sell &  0.0738 & 0.7642 &  0.6903 \\
\hline
 sole &    sell &  0.0738 & 0.7605 &  0.6866 \\
\hline
 ... &     ... &       ... &      ... &       ... \\
\hline
 eight &     eat &  0.7196 & 0.4352 & -0.2844 \\
\hline
 allege &     ledge &  0.7149 & 0.4172 & -0.2976 \\
\hline
 ache &     egg &  0.7734 & 0.4580 & -0.3154 \\
\hline
 engels &   angle &  0.8318 & 0.4986 & -0.3331 \\
\hline
 bullion &    bull &  0.7814 & 0.4128 & -0.3685 \\
 \hline 
 \\
\end{tabular}
\end{center}

Figure \ref{fig:r_pun} shows the density distribution of the cosine similarity between the test word-pairs for our proposed algorithm and PSSVec. It is observed that our distribution closely resembles a Gaussian, with smaller variance and higher mean as compared to PSSVec embeddings. 

\begin{center}
  \includegraphics[width=\linewidth]{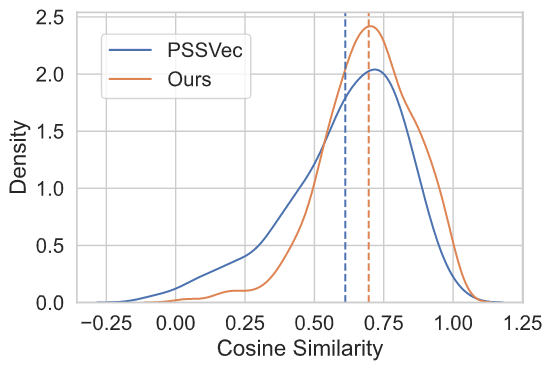}
  \captionof{figure}{Cosine similarity distribution comparison}
  \label{fig:r_pun}
\end{center}



\section{Conclusion}
\label{sec:conclusion}
In this work, we propose a method to obtain the similarity between a pair of words given their phoneme sequence and a phonetic features mapping for the language. Overcoming the short-coming of previous embedding based solutions, our algorithm can also be used to compare the similarity of new words coming on the fly, even if they are not seen in the training data.
Further, since embeddings have proven helpful in various cases, we also present a word embedding algorithm utilizing our similarity metric. Our embeddings are shown to perform better than previously reported results in the literature. We claim that our approach is generic and can be extended to any language for which phoneme sequences can be obtained for all its words. To showcase this, results are presented for $2$ languages - English and Hindi.

Going forward, we wish to apply our approach to more Indian languages, utilizing the fact that there is no sophisticated grapheme-to-phoneme conversion required in most of them as the majority of Indian languages have orthographies with a high grapheme-to-phoneme and phoneme-to-grapheme correspondence. The approach becomes more valuable for Hindi and other Indian languages as there is very less work done for them and adaptation of a generic framework allows for diversity and inclusion. We also want to take advantage of this similarity measure and create a word-based speech recognition system that can be useful for various tasks like limited vocabulary ASR, keyword spotting, and wake-word detection.

\bibliography{anthology,custom}
\bibliographystyle{acl_natbib}




\end{document}